\documentclass{esannV2}
\usepackage[]{graphicx}
\usepackage[latin1]{inputenc}
\usepackage{amssymb,amsmath,array}

\begin{document}
\title{Verifying Deep Learning-based Decisions for Facial Expression Recognition}

\author{Ines Rieger$^1$, Ren\'e Kollmann$^2$, Bettina Finzel$^2$, Dominik Seuss$^1$, Ute Schmid$^2$
%
\thanks{This project is funded by the Federal Ministry of Education and Research, grant no. 01IS18056A and 01IS18056B (TraMeExCo). We thank Sebastian Lapuschkin and Jaspar Pahl for their support.}
%
\vspace{.3cm}\\
%
1- Fraunhofer Institute for Integrated Circuits \\
Erlangen - Germany\\
%
\vspace{.1cm}\\
2- University of Bamberg - Cognitive Systems\\
Bamberg - Germany
}


\maketitle

\begin{abstract}
Neural networks with high performance can still be biased towards non-relevant features. However, reliability and robustness is especially important for high-risk fields such as clinical pain treatment. We therefore propose a verification pipeline, which consists of three steps. First, we classify facial expressions with a neural network. Next, we apply layer-wise relevance propagation to create pixel-based explanations. Finally, we quantify these visual explanations based on a bounding-box method with respect to facial regions. Although our results show that the neural network achieves state-of-the-art results, the evaluation of the visual explanations reveals that relevant facial regions may not be considered.





\end{abstract}

\section{Introduction}
\label{sect:introduction}

For healthcare professionals it might not be easy to estimate pain, when patients are not able to communicate their pain intensity. Hence, automatically detecting pain from facial expressions can support the medical decision making process.

However, as exemplified by H{\"a}gele et al. \cite{hagele_resolving_2019}, high-performing classifiers can be biased towards the training data. It is therefore crucial to make the reasons for a classification outcome transparent instead of just relying on the performance values. To indicate the influence of each pixel on the classification outcome, H{\"a}gele et al. \cite{hagele_resolving_2019} utilize layer-wise relevance propagation (LRP) \cite{bach2015pixel}, which is an explanation method that creates heatmaps for individual images. Performing a manual inspection of all heatmaps would be a tedious task. Kohlbrenner et al. \cite{kohlbrenner2019towards} propose a quantification method for evaluating visual explanations generated by LRP. Their method is based on bounding boxes containing the objects relevant to the true class.

Based on these contributions, we propose a verification pipeline in order to evaluate neural networks, which detect pain-related facial expressions. Facial expressions are divided into isolated facial movements, each being an Action Unit (AU). Certain combinations of AUs may be indicators for pain \cite{kunz_facial_2019}. We apply our trained Residual Network (ResNet) to classify these pain-relevant Action Units. Afterwards, we use LRP to generate heatmaps in order to visualize the classifier's decision. Finally, we adapt the bounding-box approach to Action Unit specific boundaries in order to quantify the quality of these heatmaps. This way we aim to assess the domain specific performance of the classifier. 

To the best of our knowledge, LRP has not yet been applied for AU detection.  Furthermore, our verification pipeline integrates a new domain-specific evaluation method to assess the quality of classifications with the help of LRP. This way we contribute to transparency for AU classification.

The paper is organized as follows: In section \ref{sect:methods} we describe the used methods, including the network architecture and LRP. In section \ref{sect:results} we describe and discuss our results based on the application of the previously defined verification pipeline. The paper concludes with a summary as well as future prospects in section \ref{sect:future-work}.

\section{Methods}
\label{sect:methods}


\subsection{Datasets and Pre-Processing}
%
%
We use the Actor Study (center view) \cite{seuss2019emotion} and Extended Cohn-Kanade (CK+) \cite{lucey2010extended} dataset for training and testing and the UNBC-McMaster Shoulder Pain (UNBC) dataset \cite{lucey2011painful} for cross-domain evaluation. The Actor Study dataset was recently introduced and will be made publicly available for non-commercial research \cite{seuss2019emotion}.
While the Actor Study and CK+ dataset provide posed facial expressions, the UNBC dataset shows in-the-wild pain expressions. We consider all frames annotated with pain-relevant Action Units.\\
%
%
%
%
%
The faces are cropped with the Sophisticated High-speed Object Recognition Engine SHORE\texttrademark \cite{kublbeck2006face}. Furthermore, SHORE\texttrademark is used to detects 68 facial keypoints according to the Multi-PIE landmark scheme \cite{gross2010multi}. All facial image crops are scaled to 112 x 112 pixels. The pixel values are normalized in a range of $[0,1]$.
Study participant with id 16 from the Actor Study dataset is omitted, since there are only 34 images containing the pain-relevant Action Units. Thus, we consider 20 participants of the Actor Study dataset with a total of 33012 images. The pre-processed CK+ dataset contains 8657 facial images. The pre-processed UNBC dataset contains 9313 facial images. Class AU27 from the UNBC dataset is omitted for evaluation since it contains only 16 images. The distribution of the training and evaluation datasets are highly imbalanced.

\subsection{Residual Network (ResNet)}
\label{subsect:resNet}


We use a parameter-reduced 18-layer ResNet \cite{rieger2019towards} based on He et al. \cite{he2016deep}, but with the ReLU activation function  for the hidden layers and the sigmoid function for the output layer. The threshold is $0.5$ to classify the Action Unit as detected or not detected. Because of the imbalanced training dataset, our loss function is a weighted binary cross-entropy. As optimizer we use the stochastic gradient descent (SGD) with a learning rate of $0.01$ and a momentum of $0.95$. 

Our model is trained on the CK+ dataset and evaluated with a leave-one-out cross validation on the Actor Study dataset.  We train each fold for 25 epochs. The best performing model of all folds is then used for further inspection by LRP (see Section \ref{subsec:resultlrp}) and cross-domain dataset evaluation (see Section \ref{subsect:resultsresnet}).
%
%
%
%
%
%
%
\subsection{Evaluating Visual Explanations}
\label{subsect:lrp}
An essential part of explaining the output of a neural network is understanding internals of its decision process. Heatmaps are a popular tool to highlight specifically interesting areas of an image. We use heatmaps generated by the LRP method to identify image sections that were prominently relevant for the network's decision \cite{bach2015pixel}. For a given input image, LRP decomposes the initial relevance given by the network's output back to its input layer - essentially assigning each pixel a relevance score that indicates how greatly it influenced the final decision. This relevance assignment is used to create heatmaps by projecting its normalized values to a predefined color spectrum. Equation \ref{eq:lrp} shows the basic decomposition rule that back-propagates relevance from one layer ($R_{k}^{(l+1)}$) to its predecessor ($R_{k}^{(l)}$), where $w_{jk}$ is the weight between two connected neurons $j$ and $k$. Variable $x_{j}$ defines the neuron activation of neuron $j$ at layer $l$. Other decomposition rules are derivatives of equation \ref{eq:lrp}.\\
\begin{equation}
R_{j}^{(l)} = \sum_{k} \frac{x_{j} w_{jk}}{\sum_{j} x_{j}w_{jk}} R_{k}^{(l+1)}
\label{eq:lrp}
\end{equation}
Action Unit detection requires close inspection of human muscle movement, therefore LRP is specifically suitable for explaining Action Unit classification due to its fine-grained, pixel-wise explanations. 
%
We adapt the best performing decomposition scheme from Kohlbrenner et al.\cite{kohlbrenner2019towards} for our model and use the implementation provided by the \textit{iNNvestigate} toolbox \cite{alber2019innvestigate}.\\
%
In order to evaluate the results of different decomposition strategies regarding their conciseness and class specific relevance distribution, Kohlbrenner et al. \cite{kohlbrenner2019towards} introduce the idea of bounding boxes as a reference to identify how well relevance is gathered around the true object. 
All (positive) relevance for an object class is supposed to be located in close proximity to the object itself. Large portions of relevance located in other parts of the image indicate that the model has not understood the problem as intended. 

For our experiments we adopt the measure of $\mu$ (equation \ref{eq:measures}), where $\mu$ describes the ratio of \textit{positive} relevance located inside the bounding box ($R_{inside}$) in comparison to the total \textit{positive} relevance distributed in the entire image  ($R_{total}$).

\begin{equation}
\mu = \frac{R_{inside}}{R_{total}}
\label{eq:measures}
\end{equation}

In our experiments we compute average $\mu$ scores for every Action Unit on every dataset. To create bounding boxes for every Action Unit we use specific facial landmarks identified by SHORE\texttrademark  as visualized in figure \ref{img:bounding_boxes}. 

\section{Results}
\label{sect:results}

\subsection{ResNet}\label{subsect:resultsresnet}
Table \ref{table:evalActorStudy} presents our averaged k-fold results of the Actor Study dataset. We compare our deep learning approach to the cross-dataset baseline with OpenFace \cite{baltruvsaitis2016openface} provided by Seuss et al. \cite{seuss2019emotion}. OpenFace \cite{baltruvsaitis2016openface} is a behavior analysis toolkit and can recognize 18 Action Units. It estimates Action Units with the help of geometric features, face alignment and masking. Our deep learning approach outperforms OpenFace \cite{baltruvsaitis2016openface} on the Actor Study dataset.

\begin{table}[h]
	\centering
	\begin{tabular}{ | l | l | l | l | l | l | l | l | l |}
		\hline
		 AUs: & 4  & 6  & 7 & 9 & 10  & 25  & 26  & 27 \\ 
		 \hline
		OpenFace \cite{seuss2019emotion} & 0.42 & 0.47 & 0.33 & 0.21 & 0.19 & 0.44 & 0.37 & - \\ 
		ResNet (\textbf{ours}) & \textbf{0.68} & \textbf{0.55} & \textbf{0.62} & \textbf{0.48} & \textbf{0.32} & \textbf{0.83} & \textbf{0.60} & \textbf{0.57} \\
		\hline
	\end{tabular}
	\caption{Evaluation on Actor Study dataset for the different pain Action Units. Results are measured in F1 Score. Best results are in bold.}
	\label{table:evalActorStudy}
\end{table}

Table \ref{table:evalUNBC} shows a cross-domain evaluation of our best performing model of all folds on the UNBC dataset, which is important to prove the generalizability of an approach. Our model performs differently well on predicting various AUs, e.g. due to the similarity of AUs or the frequency of occurrence of the AUs in the training data. To the best of our knowledge, we could find two other approaches, who did a cross-domain evaluation on the Action Units of the UNBC dataset. Peng et al. \cite{peng2018weakly} propose a weakly supervised approach with adversarial training using domain knowledge such as dependencies between Action Units. 
Tu et al. \cite{tu2019idennet} focus on identity-dependent image features, which they extract by a face clustering network. Due to the different approaches, it is difficult to compare the results, but it shows nevertheless the competitive generalization ability of our network.

\begin{table}[h]
	\centering
	\begin{tabular}{ | l | l | l | l | l | l | l | l |}
		\hline
		AUs: & 4  & 6  & 7 & 9 & 10  & 25  & 26 \\ 
		\hline
		RAN \cite{peng2018weakly} & 0.16 & 0.29 &  0.01  & \textbf{0.45} & \textbf{0.32} & - & -  \\ 
		IdenNet \cite{tu2019idennet} & 0.10 & 0.33 &  0.13  & - & 0.03 & - & -\\ 
		ResNet (\textbf{ours}) & \textbf{0.19} & \textbf{0.43} & \textbf{0.38} & 0.09 & 0.02 & \textbf{0.34} & \textbf{0.03} \\
		\hline
	\end{tabular}
	\caption{Cross-domain evaluation on UNBC dataset on the pain Action Units. Results are measured in F1 Score. Best results are in bold.}
	\label{table:evalUNBC}
\end{table}

\subsection{Visual Explanations}
\label{subsec:resultlrp}


Table \ref{tab:mu} shows the average $\mu$-values reached for every Action Unit on all three datasets. It shows that $\mu$ hardly exceeds the value of \textit{0.5}. This means that on average less than 50\% of the positive relevance is located inside the respective bounding box.







\begin{table}[h!]
\centering
\begin{tabular}{|c|c|c|c|c|c|c|}\hline
AUs & \multicolumn{2}{|c|}{Actor Study} & \multicolumn{2}{|c|}{CK+} & \multicolumn{2}{|c|}{UNBC}\\ 
\hline
& $\mu$ & n & $\mu$ & n & $\mu$ & n\\ 
\hline
4 & 0.50 & 11249 & 0.554 & 3753 & 0.50 & 291\\ 
6 & 0.42 & 6887 & 0.475 & 2174 & 0.50 & 1740\\ 
7 & 0.43 & 13098 & 0.52 & 2222 & 0.57 & 1439\\ 
9 & 0.36 & 3363 & 0.736 & 1191 & 0.69 & 213\\ 
10 & 0.20 & 3143 & 0.516 & 383 & 0.44 & 7\\ 
25 & 0.47 & 13497 & 0.35 & 5358 & 0.35 & 738\\ 
26 & 0.49 & 7537 & 0.326 & 882 & 0.20 & 33\\ 
27 & 0.49 & 1214 & 0.341 & 1285 & - & -\\ 
\hline
\end{tabular}
\caption{Average $\mu$-values for correctly predicted Action Units.}
\label{tab:mu}
\end{table}

\begin{figure}
\begin{minipage}{0.3\textwidth}
\centering
\includegraphics[width=\textwidth]{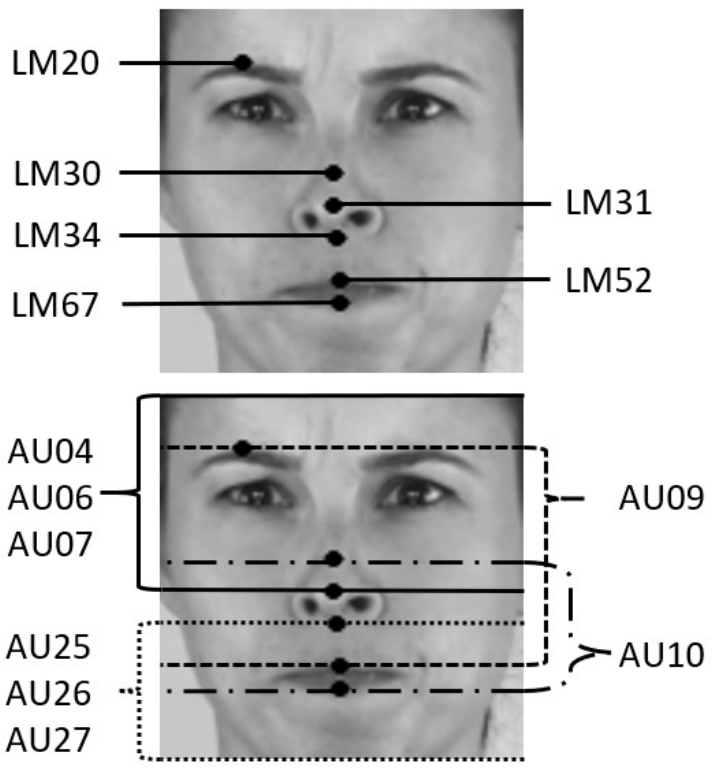}
\caption{Top: relevant landmark IDs, bottom: bounding boxes for each Action Unit}
\label{img:bounding_boxes}
\end{minipage}
\hfill
\begin{minipage}{0.68\textwidth}
\includegraphics[width=\textwidth]{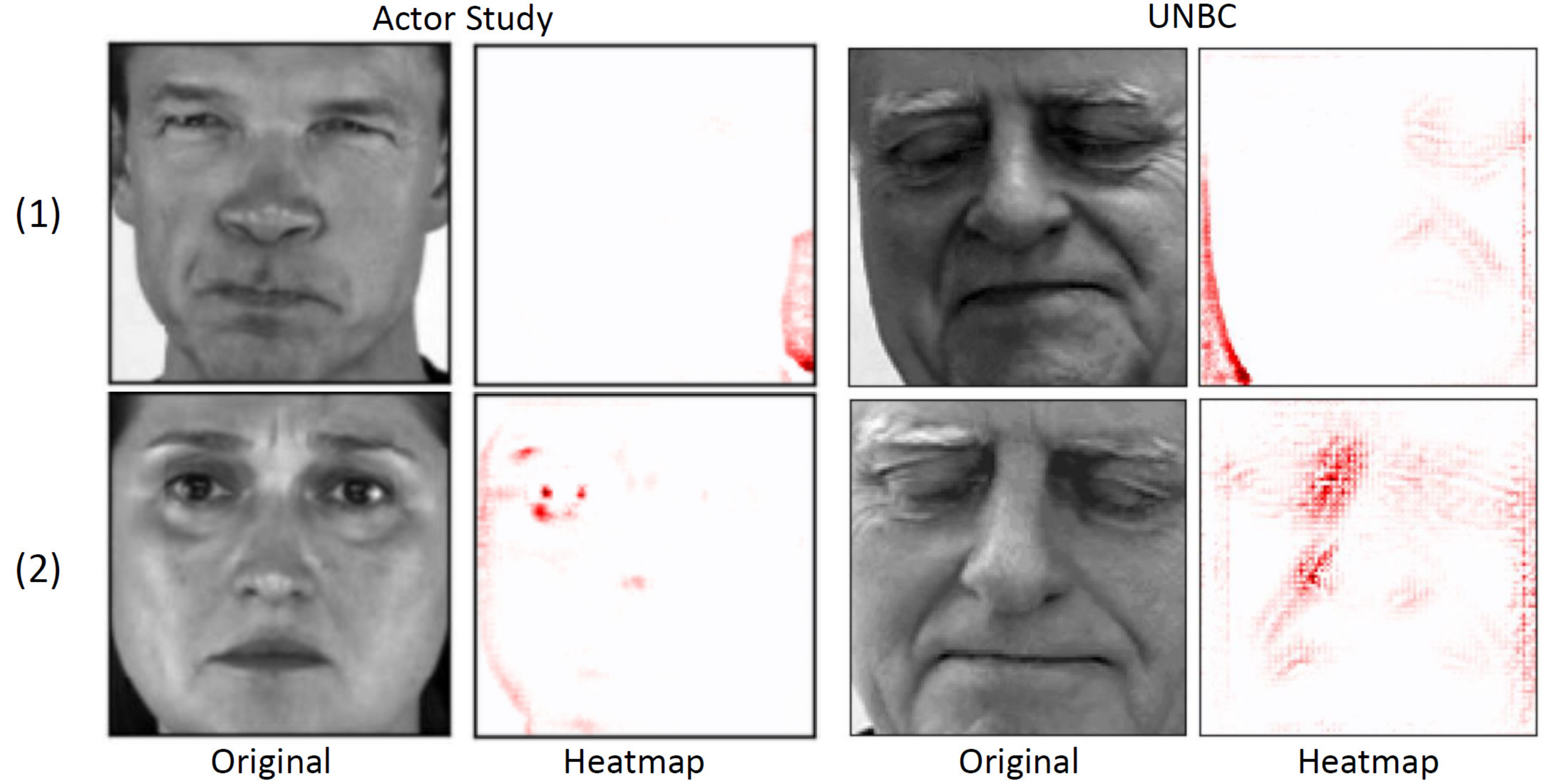}
\caption{How the neural network sees AU04 (brow lowerer) while correctly classifying it: (1) low relevance in bounding box, (2) high relevance in bounding box; left: Actor Study, right: UNBC}
\label{img:heatmaps}
\end{minipage}
\end{figure}

Figure \ref{img:heatmaps} shows four example heatmaps for a correctly classified AU04, two from the actor study dataset (training data, left) and two from the UNBC dataset (test data, right). Heatmaps like in figure \ref{img:heatmaps}(1) have $\mu$-values closer to zero and heatmaps like in figure \ref{img:heatmaps}(2) have $\mu$-values closer to one. In instances similar to figure \ref{img:heatmaps}(1), we can see that the neural network focuses on the white background in the lower part of the image. In other instances similar to figure \ref{img:heatmaps}(2), the network focuses on regions more closely connected to relevant regions. From a well performing network we expect that it identifies the background as irrelevant, but focuses on areas close to the true object. For instance, to detect AU04 (brow lowerer), we expect the network to focus on a region close to the brows. 

\subsection{Discussion}
The quantitative heatmap analysis reveals that the ResNet's decisions are not reasonable when domain knowledge is considered. This shows a discrepancy to the state-of-the-art results of the ResNet. We suspect that the reason for this discrepancy might be intrinsic bias in the input data or an overfitted neural network. Looking only at the precision of a network is thus not enough.


\section{Conclusion and Future Prospects}
\label{sect:future-work}


In this paper we propose a verification pipeline to evaluate a trained neural network. We exemplify the necessity of a quantitative analysis on the use case of pain-relevant Action Unit detection. We use LRP to visualize the networks decisions and quantify these with a domain specific bounding box approach. By using our verification pipeline we can uncover a discrepancy between the state-of-the-art results of the neural network and the quantitative analysis of the heatmaps.
We therefore show the importance of evaluating a neural network beyond precision. This is especially important for a high-risk field such as medicine, where we see our proposed verification pipeline as an essential component to build up trust for classification models.

Based on the outcome we want to further improve our verification pipeline by more domain specific measures and analyze a broader set of neural networks.



\begin{footnotesize}




\bibliographystyle{unsrt}
\bibliography{esann_bib}

\begin{thebibliography}{10}

\bibitem{hagele_resolving_2019}
Miriam H{\"a}gele~et al.
\newblock Resolving challenges in deep learning-based analyses of
  histopathological images using explanation methods.
\newblock {\em arXiv}, August 2019.

\bibitem{bach2015pixel}
Sebastian Bach~et al.
\newblock {On Pixel-Wise Explanations for Non-linear Classifier Decisions by
  Layer-Wise Relevance Propagation}.
\newblock {\em PloS one}, 10(7):e0130140, 2015.

\bibitem{kohlbrenner2019towards}
Maximilian Kohlbrenner~et al.
\newblock {Towards Best Practice in Explaining Neural Network Decisions with
  LRP}.
\newblock {\em arXiv preprint arXiv:1910.09840}, 2019.

\bibitem{kunz_facial_2019}
Miriam Kunz~et al.
\newblock Facial muscle movements encoding pain -- a systematic review.
\newblock {\em Pain}, 160(3):535--549, 2019.

\bibitem{seuss2019emotion}
Dominik Seuss~et al.
\newblock Emotion expression from different angles: A video database for facial
  expressions of actors shot by a camera array (in press).
\newblock In {\em Conf. Proc. ACII}, 2019.

\bibitem{lucey2010extended}
Patrick Lucey~et al.
\newblock {The Extended Cohn-Kanade Dataset (CK+): A complete dataset for
  action unitand emotion-specified expression}.
\newblock In {\em Conf. Proc. CVPR-Workshop}, pages 94--101. IEEE, 2010.

\bibitem{lucey2011painful}
Patrick Lucey~et al.
\newblock Painful data: The {UNBC-McMaster} shoulder pain expression archive
  database.
\newblock In {\em FG}, pages 57--64. IEEE, 2011.

\bibitem{kublbeck2006face}
Christian K{\"u}blbeck and Andreas Ernst.
\newblock Face detection and tracking in video sequences using the
  modifiedcensus transformation.
\newblock {\em Image and Vision Computing}, 24(6):564--572, 2006.

\bibitem{gross2010multi}
Ralph Gross~et al.
\newblock {Multi-PIE}.
\newblock {\em Image and Vision Computing}, 28(5):807--813, 2010.

\bibitem{rieger2019towards}
Ines Rieger~et al.
\newblock {Towards Real-Time Head Pose Estimation: Exploring Parameter-Reduced
  Residual Networks on In-the-wild Datasets}.
\newblock In {\em Conf. Proc. IEA/AIE}, pages 123--134. Springer, 2019.

\bibitem{he2016deep}
Kaiming He~et al.
\newblock {Deep Residual Learning for Image Recognition}.
\newblock In {\em Conf. Proc. CVPR}, pages 770--778, 2016.

\bibitem{alber2019innvestigate}
Maximilian Alber~et al.
\newblock {iNNvestigate Neural Networks!}
\newblock {\em Journal of Machine Learning Research}, 20(93):1--8, 2019.

\bibitem{baltruvsaitis2016openface}
Tadas Baltru{\v{s}}aitis~et al.
\newblock {OpenFace: an open source facial behavior analysis toolkit}.
\newblock In {\em Conf. Proc. WACV}, pages 1--10. IEEE, 2016.

\bibitem{peng2018weakly}
Guozhu Peng and Shangfei Wang.
\newblock {Weakly Supervised Facial Action Unit Recognition through Adversarial
  Training}.
\newblock In {\em Conf. Proc. CVPR}, pages 2188--2196, 2018.

\bibitem{tu2019idennet}
Cheng-Hao Tu~et al.
\newblock {IdenNet: Identity-Aware Facial Action Unit Detection}.
\newblock In {\em Conf. Proc. FG}, pages 1--8. IEEE, 2019.

\end{thebibliography}

\end{footnotesize}


\end{document}